\documentclass{Interspeech}

\interspeechcameraready 

\title{Leveraging Large Language Models in Visual Speech Recognition: \\ Model Scaling, Context-Aware Decoding, and Iterative Polishing}

\author[affiliation={1}]{Zehua}{Liu}
\author[affiliation={1}]{Xiaolou}{Li}
\author[affiliation={1}]{Li}{Guo}
\author[affiliation={1}]{Lantian}{Li}
\author[affiliation={2}]{Dong}{Wang}

\affiliation{School of Artificial Intelligence}{Beijing University of Posts and Telecommunications}{China}
\affiliation{Center for Speech and Language Technologies}{Tsinghua University}{China}

\email{\{lzh211,lixiaolou,lilt,guoli\}@bupt.edu.cn,\{wangdong99\}@mails.tsinghua.edu.cn}


\usepackage{comment}

\usepackage{amsmath}
\usepackage{amssymb}

\usepackage{graphicx}
\usepackage{subcaption}
\usepackage{float}

\usepackage{booktabs}
\usepackage{multirow}
\usepackage{makecell}

\begin{document}
\maketitle
\begin{abstract}

Visual Speech Recognition (VSR) transcribes speech by analyzing lip movements. Recently, Large Language Models (LLMs) have been integrated into VSR systems, leading to notable performance improvements. However, the potential of LLMs has not been extensively studied, and how to effectively utilize LLMs in VSR tasks remains unexplored. This paper systematically explores how to better leverage LLMs for VSR tasks and provides three key contributions: (1) Scaling Test: We study how the LLM size affects VSR performance, confirming a scaling law in the VSR task. (2) Context-Aware Decoding: We add contextual text to guide the LLM decoding, improving recognition accuracy. (3) Iterative Polishing: We propose iteratively refining LLM outputs, progressively reducing recognition errors. 
Extensive experiments demonstrate that by these designs, the great potential of LLMs can be largely harnessed, leading to significant VSR performance improvement.
\end{abstract}

\section{Introduction}

Visual Speech Recognition (VSR), often referred to as lip reading,
is a technique that transcribes spoken content by analyzing a speaker's lip movements~\cite{assael2016lipnet,shillingford19_interspeech,ma2023auto}.
Typically, a VSR system takes a silent video containing the speaker's lip movements as input and outputs the corresponding text.
VSR has a wide range of real-world applications, including public safety, assistance for the elderly and hearing-impaired, and deepfake video detection~\cite{li24ta_interspeech}, highlighting its significant practical value and research importance.

In recent years, substantial progress has been made to enhance the performance of VSR systems.
For example, Ma et al.~\cite{ma2021end} proposed a powerful VSR framework that involves a Conformer~\cite{gulati20_interspeech} as the encoder and
a Hybrid CTC/Transformer~\cite{petridis2018audio} as the decoder.
They also used a pre-trained Transformer-based language model~\cite{vaswani2017attention} to facilitate decoding by shallow fusion.
Other studies, such as SyncVSR~\cite{ahn24_interspeech} and AlignVSR~\cite{liu2024alignvsr},
leverage the audio modality as auxiliary information and
align the visual and audio streams based on their temporal synchronization.
Recently, some researchers turned their focus to Large Language Models (LLMs). The key idea is that the visual information is incomplete in nature, which means that VSR must rely on external information to fill up the information gap, especially the language information. LLMs can provide rich language information via their strong contextual understanding and language reasoning capabilities, thus supposedly benefiting VSR.
So far, LLMs have demonstrated significant potential in various vision-text tasks,
achieving remarkable success in video captioning~\cite{tang2023video, shu2023audio, zhou2024survey, cheng2025interactive},
image captioning~\cite{bianco2023improving, patel2025alt, tiet2025image}, and other vision-text tasks.
Inspired by these successes, researchers have explored the possibility to use LLMs in VSR.
For example, \cite{cappellazzo2024large, yeo2024visual} utilize AV-Hubert~\cite{shi2022learning} as the visual encoder and replace the traditional Transformer decoder with a pre-trained LLM. Fine-tuning is then employed to align the visual and text modalities by lightweight adaptation algorithms such as QLoRA~\cite{dettmers2024qlora}.
While these methods represent a promising direction, the research is still limited. For instance, the scaling law of the LLMs in VSR has not been well studied. Although Cappellazzo et al.~\cite{cappellazzo2024large} explored the impact of LLM scale on VSR performance using TinyLlama (1.1B) and Llama3.1-8B~\cite{dubey2024llama}, these LLMs are relatively small, leaving the potential of larger-scale LLMs unexplored.
Moreover, we argue that current research in VSR has yet to fully exploit the exceptional capabilities of LLMs in contextual understanding and reasoning. More advanced methods are required to harness their rich linguistic knowledge and powerful inference abilities.
\begin{figure*}[h]
    \centering
    \includegraphics[width=0.95\linewidth]{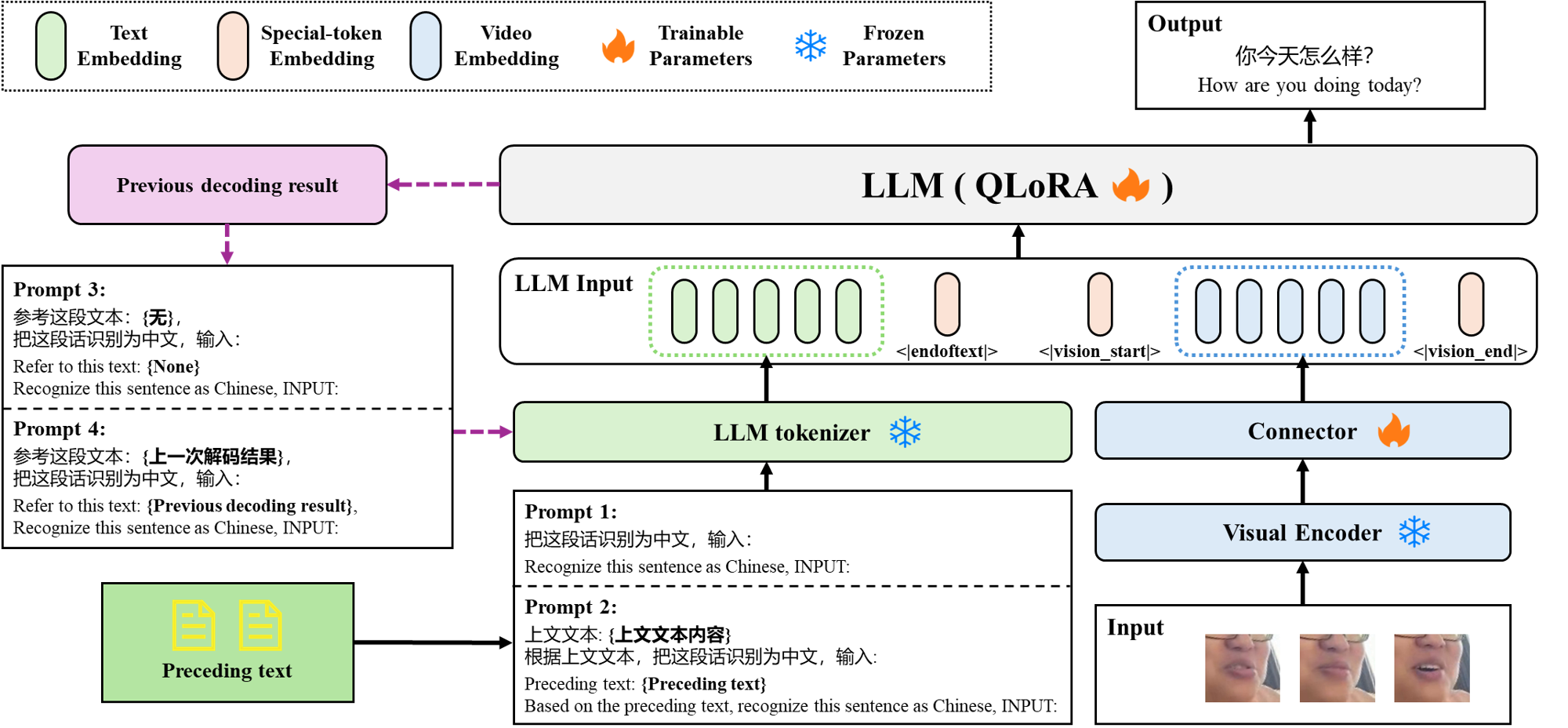}
    \caption{Our VSR-LLM architecture.}
    \label{fig:arch}
\end{figure*}

To fully utilize the potential of LLMs in VSR tasks, this paper presents a comprehensive study and provides three key contributions:
(1) \textbf{Scaling Test}: We thoroughly study how the scale of LLMs affects VSR performance and observe a clear Scaling Law—larger LLMs consistently yield better results in VSR tasks.
(2) \textbf{Context-Aware Decoding}: We introduce contextual text to guide the LLM during inference, which significantly enhances the recognition accuracy, especially when dealing with homophones and context-dependent sentences.
(3) \textbf{Iterative Polishing}: We design an iterative decoding strategy that refines the LLM's output gradually,
which further reduces recognition errors.
Our experimental results demonstrate the great potential of LLMs in advancing VSR tasks and provide new insights for future research in this domain.

The rest of this paper is organized as follows:
Section~\ref{sec:method} introduces the main architecture of our VSR-LLM system, followed by the pipelines of our proposed context-aware decoding and iterative polishing approaches.
Section~\ref{sec:setup} details the experimental setup, and Section~\ref{sec:result} presents the results and analyses.
Finally, Section~\ref{sec:conc} concludes the paper.

\section{Our VSR-LLM Architecture}
\label{sec:method}

\subsection{Main Architecture}
The architecture of our VSR-LLM system is adapted from ~\cite{yeo2024visual}, and has been illustrated in Figure~\ref{fig:arch}.
The input is a preprocessed silent video containing the mouth area of the speaker's face.
A \textbf{Visual Encoder} extracts frame-wise visual features from the input video.
The extracted visual features are then passed through a linear projection layer, referred to as the \textbf{Connector},
which aligns the dimensionality of the visual features with the LLM's embedding space, producing the video embedding.
The LLM also accepts various types of text prompts, as will be detailed shortly. The input text prompt is tokenized using a \textbf{LLM Tokenizer} and mapped into an embedding space, resulting in the text embedding.
To effectively handle multi-modal inputs, we adopt a strategy inspired by Qwen2-VL~\cite{wang2024qwen2} that concatenates the text and visual embedding vectors. 
Three special tokens are introduced to perform the concatenation:
\begin{itemize}
\item \texttt{<|endoftext|>} marks the end of the textual input; 

\item \texttt{<|vision\_start|>} denotes the start of the visual input; 

\item \texttt{<|vision\_end|>} indicates the end of the visual input.

\end{itemize}
These special tokens explicitly delineate textual and visual modalities.
The concatenated embeddings are then fed into LLMs for training and inference. 
The main text prompt is \textbf{Prompt 1}, which specifies the task in the form ``Recognize the sentence as Chinese, INPUT:".

For the baseline model, a cross-entropy loss function $\mathcal{L}_{\text{base}}$ is employed for model training, given by:
\begin{align}
\mathcal{L_{\text{base}}} = & - \sum_{l=1}^{L} \log p(y^l | V, y^{<l})
\end{align}
\noindent where \(V\) represents the input video, \(y^l\) denotes the \(l^{th}\) token of the ground truth transcription, 
\(y^{<l}\) is the sequence of preceding generated tokens, and \(L\) is the total number of tokens of the ground truth transcription.

The LLM parameters are fine-tuned using the QLoRA algorithm~\cite{dettmers2024qlora}, which enables efficient adaptation of large-scale language models through low-rank parameter updates.

\subsection{Context-Aware Decoding}

A key advantage of LLMs is their capability of taking into account long-range contextual information, and this has been utilized by a couple of studies~\cite{bai2024seed,chen2024salm,lakomkin2024end} in speech recognition, where related texts are involved in the LLM input as extra context. 
The related texts could be generated by an LLM or other annotations, e.g., topics or keywords. 

Motivated by these studies, we design a \emph{context-aware decoding} strategy in our VSR-LLM framework.
Specifically, we use the text transcription (produced by speech recognition) of the preceding 30-second video segment as the input of the LLM, specified by \textbf{Prompt 2} in Fig.~\ref{fig:arch}.
We argue that incorporating the contextual text enables the LLM to take long-range contexts into account, thus providing extra information when decoding the present utterance. 
We also expect that this approach is especially effective in disambiguating homophones and recognizing context-dependent phrases. 
For instance, visual cues that correspond to multiple possible transcriptions (e.g., homophones) can be more accurately resolved with the additional context.

The training objective for the context-aware decoding is defined as follows:
\begin{align}
\mathcal{L_{\text{context}}} = & - \sum_{l=1}^{L} \log p(y^l | V,T, y^{<l})
\end{align}
\noindent where \(T\) represents the context, i.e., the preceding text in our case.

\subsection{Iterative Polishing}

Current VSR systems typically employ a single-pass decoding, where the model generates the transcription in one forward pass.
However, such a single-pass decoding may lead to suboptimal outputs, as the decoding errors have no chance of being corrected~\cite{zhu2024eliciting}.
To address this issue, we propose an \emph{iterative polishing} aproach that allows the model to refine its outputs over multiple decoding iterations.

As shown in Figure~\ref{fig:arch}, during training, the initial transcription is generated using \textbf{Prompt 3}.
This preliminary output is then fed back into the model as auxiliary information to construct \textbf{Prompt 4}, 
which asks the LLM to refine the initial recognition.
Through this iterative process, the model learns self-correction and progressively improves its decoding.

To make the model support iterative polishing, the training process needs a slight change. 
Firstly, we need to run the baseline model to produce the initial result $P$, and then construct the training samples by constructing triplets $\{V,P,Y\}$. The objective function is formulated as:
\begin{align*}
\mathcal{L_{\text{iter}}} = & - \sum_{l=1}^{L} \log p(y^l | V,P, y^{<l}) \\
& - \sum_{l=1}^{L} \log p(y^l | V, y^{<l}) 
\tag{3}
\end{align*}
\noindent where \(P\) denotes the transcription generated from the first decoding. 

During inference, the model first generates an initial transcription and then refines it through multiple iterations, each time using the previous output as an additional context. 
This process continues until the decoding results in convergence or a predefined maximum number of iterations is reached.

\subsection{Context-aware Iterative Decoding}
\label{sec:caid}

To fully leverage the benefits of both context-aware decoding and iterative polishing,
we further propose a unified strategy termed \emph{context-aware iterative decoding}.
This method combines the advantages of contextual information with iterative polishing, further improving the recognition accuracy. 

In the first iteration, the prompt is constructed as follows:

\texttt{Preceding text: \{Preceding text\}. Based on the preceding text and the reference: \{None\}, recognize this sentence as Chinese, INPUT:}

For the subsequent iterations, the prompt takes into account the decoding results of the previous iteration:

\texttt{Preceding text: \{Preceding text\}. Based on the preceding text and the reference: \{Previous decoding result\}, recognize this sentence as Chinese, INPUT:}

The corresponding training objective for context-aware iterative decoding is:
\begin{align*}
\mathcal{L_{\text{context\_iter}}} = & - \sum_{l=1}^{L} \log p(y^l | V,P,T, y^{<l}) \\
& - \sum_{l=1}^{L} \log p(y^l | V,T, y^{<l}) 
\tag{4}
\end{align*}
\noindent where \(T\) denotes the preceding text, \(P\) is the result from the previous decoding iteration.

During inference, the model leverages both preceding context and iterative decoding. 
It refines its outputs by several iterations, using each iteration's result as input for the next.  
The preceding context is presumed to be available\footnote{Our experiments showed that there is not much difference for various types of contexts, as long as it contains sufficient indicating words for the topic.}. 
We will show that this combination offers further performance improvement.

\section{Experiment Setup}
\label{sec:setup}

\subsection{Data}
To evaluate the effectiveness of the proposed methods, we need to select an appropriate dataset that enables LLMs to demonstrate their potential.  
Popular VSR datasets like LRS2 and LRS3~\cite{afouras2018deep} contain relatively short video clips (usually under 10 seconds),  
limiting the availability of contextual information and thus not suitable for demonstrating LLMs' ability in context learning.  We finally chose the CNVSRC.Single dataset~\cite{chen2024cnvsrc}, which provides richer context.

CNVSRC.Single was released as part of the CNVSRC 2023 challenge and contains single-speaker data.  
The dataset comprises a training set containing 83 hours of video from 23,064 video clips, a validation set with 10 hours of video from 2,882 video clips,  
and a test set with 10 hours of video from 2,881 video clips.  
Each video clip (average length of 13 seconds) is extracted from long videos (averaging 15 minutes) collected from the internet.  
Additionally, the topics of these videos are primarily focused on real-time political news, offering strong domain specificity and contextual coherence. 
These features make CNVSRC.Single a suitable dataset to train and test context-aware models.

For experiments involving \emph{context-aware decoding}, we extracted the preceding text for each video clip.  
Specifically, a 30-second audio segment preceding each video clip was retrieved from the raw video and  
was transcribed into text using the ASR system recommended by CN-CVS~\cite{chen2023cn}\footnote{\url{https://github.com/PaddlePaddle/PaddleSpeech}}.  
The transcribed audio serves as the preceding text, ranging from 2 to 593 Chinese characters, with an average of 182 characters.

\subsection{Preprocessing}

For data preprocessing, we followed the pipeline of the CNVSRC 2023 baseline~\cite{chen2024cnvsrc}.  
Specifically, RetinaFace~\cite{deng_retinaface_2019} was firstly applied to detect the facial region in each video frame. Then, the FAN model~\cite{bulat_how_2017} was used to extract facial landmarks, and each detected face was aligned to a standard reference using these landmarks. Finally, a $96 \times 96$ region focusing on the lips was cropped from each aligned frame, which was then used as the input to the model.

\subsection{Model}

For the visual encoder, we employed the pre-trained baseline model provided by the CNVSRC 2024 challenge for the single track\footnote{\url{http://cnceleb.org/competition}}.  
The architecture comprises a ResNet18 backbone for initial feature extraction, a 3D-CNN layer to capture short-term spatiotemporal features from lip movements, 8 2D-CNN layers for enhanced spatial feature extraction, and 12 Conformer layers to model long-range temporal dependencies and extract the final visual embeddings.

For the decoder, we compared different models: a 6-layer Transformer as the baseline and various LLMs, including  
Qwen2.5-7B, Qwen2.5-14B, Qwen2.5-32B, and Qwen2.5-32B-Instruct (an instruction-tuned version of Qwen2.5-32B).  
For detailed information on the Qwen LLM family, please refer to the official documentation~\cite{yang2024qwen2}.

\subsection{Training and Evaluation}

All experiments were trained for 17,000 steps using the Adam optimizer with hyperparameters $\beta_1 = 0.9$ and $\beta_2 = 0.98$.  
Beam search was applied during decoding with a beam width of 5, a length penalty of 0, and a repetition penalty of 6.

For experiments involving \emph{iterative polishing} and \emph{context-aware iterative decoding}, the number of iterations was set to 1 in the training phase, as ground truth is available during training, and 4 in the test phase to allow for progressive refinement.

We used Character Error Rate (CER) as the primary evaluation metric to assess model performance. 
All datasets, code, and experimental results are publicly available\footnote{\url{https://github.com/liu12366262626/VSR-LLM}}.

\section{Experimental Results}
\label{sec:result}

\subsection{Results on Scaling Test}

We first conduct a \textbf{Scaling Test} to investigate how the LLM scale affects VSR performance. 
The input prompt is in the form of \textbf{Prompt 1} in Figure~\ref{fig:arch}. 
The experimental results are presented in Table~\ref{tab:scale}.

Firstly, we can see that LLMs significantly outperform the Transformer decoder baseline, 
highlighting the advantage of employing the strong knowledge distilled from the LLMs. 
This observation is consistent with previous studies~\cite{cappellazzo2024large,yeo2024visual}.
Furthermore, as the scale of the LLM increases, VSR performance consistently improves, confirming the scaling of the LLMs in the VSR task.

\begin{table}[ht]
  \caption{CER results with different LLM decoders on CNVSRC.Single.Valid/Test sets.}
  \label{tab:scale}
  \centering
  \resizebox{0.65\linewidth}{!}{ 
  \begin{tabular}{l|c|c}
    \toprule
    \textbf{Decoder}              & \textbf{Valid}   & \textbf{Test} \\
    \midrule
    Transformer                   & 50.53\% & 49.32\% \\
    \midrule
    Qwen2.5-7B                    & 46.48\% & 45.20\% \\
    Qwen2.5-14B                   & 44.30\% & 42.94\% \\
    Qwen2.5-32B                   & 43.87\% & 42.50\% \\
    Qwen2.5-32B-Instruct          & 43.23\% & 41.86\% \\
    \bottomrule
  \end{tabular}}
\end{table}

\subsection{Results on Context-Aware Decoding}

Next, we evaluate the \textbf{Context-Aware Decoding (CAD)} strategy to test the in-context learning ability of LLMs in VSR tasks. 
Specifically, preceding textual information is provided in the prompt to guide the LLM during decoding.
The input prompt follows the form of \textbf{Prompt 2} in Figure~\ref{fig:arch}. 
The results are shown in Table~\ref{tab:context}.

It can be seen that adding the preceding text leads to a significant reduction in CER,
indicating that LLMs can effectively leverage the in-context information to infer the speech content. 
In addition, the improvements are more pronounced with the larger Qwen2.5-32B-Instruct model compared to the smaller Qwen2.5-7B.
This aligns with expectations that larger LLMs have stronger contextual understanding and reasoning capabilities.

Moreover, through careful post-analysis, we observed that context-aware decoding effectively leverages in-context learning to handle challenging cases, 
such as homophones and domain-specific terminology. 
Some of these cases are publicly available.
This confirms the hypothesis that LLMs, by using context information, can tackle ambiguous scenarios where visual information alone may be insufficient.

\begin{table}[ht]
  \caption{CER results with Context-Aware Decoding (CAD) on CNVSRC.Single.Valid/Test sets.}
  \label{tab:context}
  \centering
  \resizebox{0.65\linewidth}{!}{ 
  \begin{tabular}{l|c|c}
    \toprule
    \textbf{Decoder}              & \textbf{Valid}   & \textbf{Test} \\
    \midrule
    Qwen2.5-7B                    & 46.48\% & 45.20\% \\
    ~~ + CAD                      & 44.58\% & 43.20\% \\
    \midrule
    Qwen2.5-32B-Instruct          & 43.23\% & 41.86\% \\
    ~~ + CAD                      & 40.69\% & 38.94\% \\
    \bottomrule
  \end{tabular}}
\end{table}

\subsection{Results on Iterative Polishing}

We then evaluate the impact of \textbf{Iterative Polishing (IP)} to assess whether the multi-round decoding can improve VSR performance 
by leveraging the LLM's self-correction abilities. 
The prompts are in the form of \textbf{Prompt 3} and \textbf{Prompt 4} in Figure~\ref{fig:arch}. 
The results are summarized in Table~\ref{tab:iter}.

The results show consistent improvements in CER after iterative polishing, though the gains are relatively limited. 
This could be due to the fact that the LLM models are not extensively fine-tuned on the refinement task. 

\begin{table}[ht]
  \caption{CER results with Iterative Polishing (IP) on CNVSRC.Single.Valid/Test sets.}
  \label{tab:iter}
  \centering
  \resizebox{0.65\linewidth}{!}{ 
  \begin{tabular}{l|c|c}
    \toprule
    \textbf{Decoder}              & \textbf{Valid}   & \textbf{Test} \\
    \midrule
    Qwen2.5-7B                    & 46.48\% & 45.20\% \\
    ~~ + IP                       & 45.80\% & 44.75\% \\
    \midrule
    Qwen2.5-32B-Instruct          & 43.23\% & 41.86\% \\
    ~~ + IP                       & 42.83\% & 41.35\% \\
    \bottomrule
  \end{tabular}}
\end{table}

\subsection{Results on Context-Aware Iterative Decoding}

Finally, we combine the context-aware decoding and iterative polishing strategies into 
a unified \textbf{Context-Aware Iterative Decoding (CAID)} framework. 
The corresponding prompts are presented in Section~\ref{sec:caid}, and the experimental results are shown in Table~\ref{tab:fusion}.

Compared to the results in Table~\ref{tab:context} and Table~\ref{tab:iter}, the combination of both strategies leads to significant performance improvements, demonstrating that CAD and IP are complementary. 
By leveraging both in-context learning and iterative polishing, LLMs can more effectively boost VSR performance.

\begin{table}[ht]
  \caption{CER results with Context-Aware Iterative Decoding (CAID) on CNVSRC.Single.Valid/Test sets.}
  \label{tab:fusion}
  \centering
  \resizebox{0.65\linewidth}{!}{ 
  \begin{tabular}{l|c|c}
    \toprule
    \textbf{Decoder}              & \textbf{Valid}   & \textbf{Test} \\
    \midrule
    Qwen2.5-7B                    & 46.48\% & 45.20\% \\
    ~~ + CAID                     & 44.29\% & 42.87\% \\
    \midrule
    Qwen2.5-32B-Instruct          & 43.23\% & 41.86\% \\
    ~~ + CAID                     & 39.90\% & 38.18\% \\
    \bottomrule
  \end{tabular}}
\end{table}

\section{Conclusion}
\label{sec:conc}

In this paper, we explored the integration of Large Language Models (LLMs) into Visual Speech Recognition (VSR). By a comprehensive experimental study, several interesting conclusions can be drawn. First, we confirmed the existence of the LLM's scaling law in VSR tasks, demonstrating that larger LLMs consistently yield better recognition performance.
Second, we introduced two key decoding strategies to the LLM-augmented VSR framework:
The first one is Context-Aware Decoding. By adding preceding textual information to the prompt, LLMs can utilize global contextual cues, leading to more accurate transcriptions, particularly for homophones and specialized terms.
The second one is Iterative Polishing. By employing an iterative inference strategy, the LLM can progressively refine its decoding outputs, effectively self-correcting the errors of the first decoding, offering overall performance improvement.
Finally, we combined the context-aware decoding and iterative polishing strategies, leading to an even stronger VSR performance.
Our findings demonstrate that effectively leveraging LLMs can significantly improve VSR performance, providing a promising direction for future research in this field.

Looking ahead, we believe LLMs hold great potential for advancing VSR as well as its related tasks, such as VTS (visual speech synthesis).  
There are lots of studies for the future: evaluating the generalization capabilities of LLMs on larger and more diverse VSR datasets, designing a more efficient and scalable VSR-LLM framework, prompt tuning for LLMs used in VSR tasks, and effective ways to construct context information.


\bibliographystyle{IEEEtran}
\bibliography{mybib}

\begin{thebibliography}{10}
\providecommand{\url}[1]{#1}
\csname url@samestyle\endcsname
\providecommand{\newblock}{\relax}
\providecommand{\bibinfo}[2]{#2}
\providecommand{\BIBentrySTDinterwordspacing}{\spaceskip=0pt\relax}
\providecommand{\BIBentryALTinterwordstretchfactor}{4}
\providecommand{\BIBentryALTinterwordspacing}{\spaceskip=\fontdimen2\font plus
\BIBentryALTinterwordstretchfactor\fontdimen3\font minus \fontdimen4\font\relax}
\providecommand{\BIBforeignlanguage}[2]{{%
\expandafter\ifx\csname l@#1\endcsname\relax
\typeout{** WARNING: IEEEtran.bst: No hyphenation pattern has been}%
\typeout{** loaded for the language `#1'. Using the pattern for}%
\typeout{** the default language instead.}%
\else
\language=\csname l@#1\endcsname
\fi
#2}}
\providecommand{\BIBdecl}{\relax}
\BIBdecl

\bibitem{assael2016lipnet}
Y.~M. Assael, B.~Shillingford, S.~Whiteson, and N.~De~Freitas, ``{LipNet}: Sentence-level lipreading,'' \emph{arXiv preprint arXiv:1611.01599}, 2016.

\bibitem{shillingford19_interspeech}
B.~Shillingford, Y.~Assael, M.~W. Hoffman, T.~Paine, C.~Hughes, U.~Prabhu, H.~Liao, H.~Sak, K.~Rao, L.~Bennett, M.~Mulville, M.~Denil, B.~Coppin, B.~Laurie, A.~Senior, and N.~de~Freitas, ``Large-scale visual speech recognition,'' in \emph{INTERSPEECH}, 2019, pp. 4135--4139.

\bibitem{ma2023auto}
P.~Ma, A.~Haliassos, A.~Fernandez-Lopez, H.~Chen, S.~Petridis, and M.~Pantic, ``{Auto-AVSR}: Audio-visual speech recognition with automatic labels,'' in \emph{ICASSP 2023-2023 IEEE International Conference on Acoustics, Speech and Signal Processing (ICASSP)}.\hskip 1em plus 0.5em minus 0.4em\relax IEEE, 2023, pp. 1--5.

\bibitem{li24ta_interspeech}
X.~Li, Z.~Liu, C.~Chen, L.~Li, L.~Guo, and D.~Wang, ``Zero-shot fake video detection by audio-visual consistency,'' in \emph{INTERSPEECH}, 2024, pp. 2935--2939.

\bibitem{ma2021end}
P.~Ma, S.~Petridis, and M.~Pantic, ``End-to-end audio-visual speech recognition with conformers,'' in \emph{ICASSP 2021-2021 IEEE International Conference on Acoustics, Speech and Signal Processing (ICASSP)}.\hskip 1em plus 0.5em minus 0.4em\relax IEEE, 2021, pp. 7613--7617.

\bibitem{gulati20_interspeech}
A.~Gulati, J.~Qin, C.-C. Chiu, N.~Parmar, Y.~Zhang, J.~Yu, W.~Han, S.~Wang, Z.~Zhang, Y.~Wu, and R.~Pang, ``Conformer: Convolution-augmented transformer for speech recognition,'' in \emph{INTERSPEECH}, 2020, pp. 5036--5040.

\bibitem{petridis2018audio}
S.~Petridis, T.~Stafylakis, P.~Ma, G.~Tzimiropoulos, and M.~Pantic, ``Audio-visual speech recognition with a hybrid {CTC}/attention architecture,'' in \emph{2018 IEEE Spoken Language Technology Workshop (SLT)}.\hskip 1em plus 0.5em minus 0.4em\relax IEEE, 2018, pp. 513--520.

\bibitem{vaswani2017attention}
A.~Vaswani, N.~Shazeer, N.~Parmar, J.~Uszkoreit, L.~Jones, A.~N. Gomez, {\L}.~Kaiser, and I.~Polosukhin, ``Attention is all you need,'' \emph{Advances in Neural Information Processing Systems}, 2017.

\bibitem{ahn24_interspeech}
Y.~J. Ahn, J.~Park, S.~Park, J.~Choi, and K.-E. Kim, ``{SyncVSR}: Data-efficient visual speech recognition with end-to-end crossmodal audio token synchronization,'' in \emph{INTERSPEECH}, 2024, pp. 867--871.

\bibitem{liu2024alignvsr}
Z.~Liu, X.~Li, C.~Chen, L.~Guo, L.~Li, and D.~Wang, ``{AlignVSR}: Audio-visual cross-modal alignment for visual speech recognition,'' \emph{arXiv preprint arXiv:2410.16438}, 2024.

\bibitem{tang2023video}
Y.~Tang, J.~Bi, S.~Xu, L.~Song, S.~Liang, T.~Wang, D.~Zhang, J.~An, J.~Lin, R.~Zhu \emph{et~al.}, ``Video understanding with large language models: A survey,'' \emph{arXiv preprint arXiv:2312.17432}, 2023.

\bibitem{shu2023audio}
F.~Shu, L.~Zhang, H.~Jiang, and C.~Xie, ``Audio-visual {LLM} for video understanding,'' \emph{arXiv preprint arXiv:2312.06720}, 2023.

\bibitem{zhou2024survey}
P.~Zhou, L.~Wang, Z.~Liu, Y.~Hao, P.~Hui, S.~Tarkoma, and J.~Kangasharju, ``A survey on generative {AI} and {LLM} for video generation, understanding, and streaming,'' \emph{arXiv preprint arXiv:2404.16038}, 2024.

\bibitem{cheng2025interactive}
Y.-T. Cheng, J.~Wu, Z.~Ma, J.~He, X.-Y. Wei, and C.-W. Ngo, ``Interactive video search with multi-modal {LLM} video captioning,'' in \emph{International Conference on Multimedia Modeling}.\hskip 1em plus 0.5em minus 0.4em\relax Springer, 2025, pp. 302--309.

\bibitem{bianco2023improving}
S.~Bianco, L.~Celona, M.~Donzella, and P.~Napoletano, ``Improving image captioning descriptiveness by ranking and {LLM}-based fusion,'' \emph{arXiv preprint arXiv:2306.11593}, 2023.

\bibitem{patel2025alt}
V.~Patel, A.~Modi, H.~Mistry, A.~Mishra, R.~Upadhyay, and A.~Shah, ``From {Alt-text} to real context: Revolutionizing image captioning using the potential of {LLM},'' \emph{IJS-CSEIT}, vol.~11, no.~1, pp. 379--387, 2025.

\bibitem{tiet2025image}
N.~Tiet, ``Image captioning using multimodal {LLMs},'' \emph{LU-CS-EX}, 2025.

\bibitem{cappellazzo2024large}
U.~Cappellazzo, M.~Kim, H.~Chen, P.~Ma, S.~Petridis, D.~Falavigna, A.~Brutti, and M.~Pantic, ``Large language models are strong audio-visual speech recognition learners,'' \emph{arXiv preprint arXiv:2409.12319}, 2024.

\bibitem{yeo2024visual}
J.~H. Yeo, S.~Han, M.~Kim, and Y.~M. Ro, ``Where visual speech meets language: {VSP-LLM} framework for efficient and context-aware visual speech processing,'' in \emph{EMNLP}, 2024, pp. 11\,391--11\,406.

\bibitem{shi2022learning}
B.~Shi, W.-N. Hsu, K.~Lakhotia, and A.~Mohamed, ``Learning audio-visual speech representation by masked multimodal cluster prediction,'' in \emph{International Conference on Learning Representations}, 2022.

\bibitem{dettmers2024qlora}
T.~Dettmers, A.~Pagnoni, A.~Holtzman, and L.~Zettlemoyer, ``{QLoRA}: efficient finetuning of quantized {LLMs},'' \emph{Advances in Neural Information Processing Systems}, vol.~36, 2024.

\bibitem{dubey2024llama}
A.~Dubey, A.~Jauhri, A.~Pandey, A.~Kadian, A.~Al-Dahle, A.~Letman, A.~Mathur, A.~Schelten, A.~Yang, A.~Fan \emph{et~al.}, ``The {Llama} 3 herd of models,'' \emph{arXiv preprint arXiv:2407.21783}, 2024.

\bibitem{wang2024qwen2}
P.~Wang, S.~Bai, S.~Tan, S.~Wang, Z.~Fan, J.~Bai, K.~Chen, X.~Liu, J.~Wang, W.~Ge \emph{et~al.}, ``Qwen2-vl: Enhancing vision-language model's perception of the world at any resolution,'' \emph{arXiv preprint arXiv:2409.12191}, 2024.

\bibitem{bai2024seed}
Y.~Bai, J.~Chen, J.~Chen, W.~Chen, Z.~Chen, C.~Ding, L.~Dong, Q.~Dong, Y.~Du, K.~Gao \emph{et~al.}, ``{Seed-ASR}: Understanding diverse speech and contexts with {LLM}-based speech recognition,'' \emph{arXiv preprint arXiv:2407.04675}, 2024.

\bibitem{chen2024salm}
Z.~Chen, H.~Huang, A.~Andrusenko, O.~Hrinchuk, K.~C. Puvvada, J.~Li, S.~Ghosh, J.~Balam, and B.~Ginsburg, ``{SALM}: Speech-augmented language model with in-context learning for speech recognition and translation,'' in \emph{ICASSP 2024-2024 IEEE International Conference on Acoustics, Speech and Signal Processing (ICASSP)}.\hskip 1em plus 0.5em minus 0.4em\relax IEEE, 2024, pp. 13\,521--13\,525.

\bibitem{lakomkin2024end}
E.~Lakomkin, C.~Wu, Y.~Fathullah, O.~Kalinli, M.~L. Seltzer, and C.~Fuegen, ``End-to-end speech recognition contextualization with large language models,'' in \emph{ICASSP 2024-2024 IEEE International Conference on Acoustics, Speech and Signal Processing (ICASSP)}.\hskip 1em plus 0.5em minus 0.4em\relax IEEE, 2024, pp. 12\,406--12\,410.

\bibitem{zhu2024eliciting}
J.-Q. Zhu and T.~L. Griffiths, ``Eliciting the priors of large language models using iterated in-context learning,'' \emph{arXiv preprint arXiv:2406.01860}, 2024.

\bibitem{afouras2018deep}
T.~Afouras, J.~S. Chung, A.~Senior, O.~Vinyals, and A.~Zisserman, ``Deep audio-visual speech recognition,'' \emph{IEEE transactions on pattern analysis and machine intelligence}, vol.~44, no.~12, pp. 8717--8727, 2018.

\bibitem{chen2024cnvsrc}
C.~Chen, Z.~Liu, X.~Li, L.~Li, and D.~Wang, ``{CNVSRC} 2023: The first {C}hinese continuous visual speech recognition challenge,'' in \emph{INTERSPEECH}, 2024, pp. 1930--1934.

\bibitem{chen2023cn}
C.~Chen, D.~Wang, and T.~F. Zheng, ``{CN-CVS}: A {M}andarin audio-visual dataset for large vocabulary continuous visual to speech synthesis,'' in \emph{ICASSP 2023-2023 IEEE International Conference on Acoustics, Speech and Signal Processing (ICASSP)}.\hskip 1em plus 0.5em minus 0.4em\relax IEEE, 2023, pp. 1--5.

\bibitem{deng_retinaface_2019}
J.~Deng, J.~Guo, E.~Ververas, I.~Kotsia, and S.~Zafeiriou, ``{Retinaface}: Single-shot multi-level face localisation in the wild,'' in \emph{Proceedings of the IEEE/CVF conference on computer vision and pattern recognition}, 2020, pp. 5203--5212.

\bibitem{bulat_how_2017}
A.~Bulat and G.~Tzimiropoulos, ``How far are we from solving the 2{D} \& 3{D} face alignment problem?(and a dataset of 230,000 3{D} facial landmarks),'' in \emph{Proceedings of the IEEE international conference on computer vision}, 2017, pp. 1021--1030.

\bibitem{yang2024qwen2}
A.~Yang, B.~Yang, B.~Zhang, B.~Hui, B.~Zheng, B.~Yu, C.~Li, D.~Liu, F.~Huang, H.~Wei \emph{et~al.}, ``Qwen2. 5 technical report,'' \emph{arXiv preprint arXiv:2412.15115}, 2024.

\end{thebibliography}
\end{document}